\documentclass{article}
\usepackage[preprint]{neurips_2026}

\usepackage[utf8]{inputenc} % allow utf-8 input
\usepackage[T1]{fontenc}    % use 8-bit T1 fonts
\usepackage{hyperref}       % hyperlinks
\usepackage{url}            % simple URL typesetting
\usepackage{booktabs}       % professional-quality tables
\usepackage{amsfonts}       % blackboard math symbols
\usepackage{nicefrac}       % compact symbols for 1/2, etc.
\usepackage{microtype}      % microtypography
\usepackage{xcolor}         % colors
\usepackage{amsmath,amssymb,amsthm}
\usepackage{booktabs}
\usepackage{array}
\usepackage{multirow}
\usepackage{makecell}
\usepackage{graphicx}
\usepackage{tikz}
\usepackage{pgfplots}
\pgfplotsset{compat=1.18}
\usetikzlibrary{arrows.meta, positioning, shapes.geometric, calc,
                 patterns, decorations.pathreplacing}
\usepackage{algorithm}
\usepackage{algpseudocode}
\usepackage{cleveref}
\usepackage{enumitem}
\setlist{nosep,leftmargin=*}

\title{Does Synthetic Data Help? Empirical Evidence from Deep Learning Time Series Forecasters}

\newcommand{\R}{\mathbb{R}}

\newcommand{\Var}{\mathrm{Var}}
\newcommand{\Normal}{\mathcal{N}}
\newcommand{\Dirichlet}{\mathrm{Dir}}
\newcommand{\Beta}{\mathrm{Beta}}
\newcommand{\TN}{\mathrm{TN}}
\newcommand{\Dcal}{\mathcal{D}}

\author{%
  Hugo Cazaux \\
  Department of Engineering\\
  Reykjavík University\\
  Iceland, Menntavegur 1, 102 \\
  \texttt{hugot@ru.is} \\
  \And
  Eyjólfur Ingi Ásgeirsson \\
  Department of Engineering \\
  Reykjavík University \\
  Iceland, Menntavegur 1, 102 \\
  \And
  Hlynur Stefánsson \\
  Department of Engineering \\
  Reykjavík University\\
  Iceland, Menntavegur 1, 102 \\
}

\date{}

\begin{document}
\maketitle

\begin{abstract}
Synthetic data has transformed language model training, yet its role in time series forecasting remains poorly understood. We present a large-scale empirical study: nine experiment groups, 4,218 runs systematically evaluating synthetic time series augmentation across five architectures, four synthetic signals and seven datasets. The effect is sharply architecture-conditional: channel-mixing models (TimesNet, iTransformer) benefit in the majority of trials, while channel-independent models (DLinear, PatchTST) are consistently degraded. In selected low-resource settings the gains are striking: TimesNet trained on only 10\% of Weather data with synthetic augmentation surpasses the full-data baseline (4 of 16 sparsity-dataset combinations). Averaged across all architectures, augmentation hurts in 67\% of trials. We further find that only the Seasonal-Trend generator reliably helps across the tested benchmarks, and that hard curriculum switching is actively harmful (+24\% MSE degradation). These results provide concrete, actionable guidelines on how to use synthetic data: use synthetic augmentation with channel-mixing architectures, use gradual annealing schedules, and treat low-resource augmentation as architecture- and dataset-dependent. Code is available at \href{https://github.com/hugoiscracked/synthetic-ts/tree/main}{synthetic-ts}.
\end{abstract}

\section{Introduction}
\label{sec:intro}
The training of time series predictors generally respects the train-validation-test split: 80\% of the data is used for training, 10\% for validation and 10\% for testing. With the emergence of foundational models for time series forecasting, there is an increasing need for time series data. As opposed to text-based data, which can be scraped, time series often sit idle on private servers or remain locked behind paywalls. Building a large, diverse and representative dataset is a challenge, and large language models (LLM) have taught us the key importance of data in learning. Synthetic data has long been a part of the training pipeline in LLM, and the high similarity between the state-of-the-art models for language and time series hints that synthetic data should also be part of the training pipeline for forecasters. 

Synthetic time series offer several advantages over real data: unlimited supply, direct sampling from the generator (saving storage and GPU throughput), fully controllable curricula, and no compounding error from auto-regressive data pipelines. These benefits come with a critical caveat: if the synthetic distribution does not align with the target domain, augmentation can degrade rather than improve performance. Understanding how to correctly use synthetic data is what this paper directly addresses.

In this article, we propose a thorough overview of the integration of synthetic data in the training of time series forecasters. We start by creating four difficulty-conditioned bundles representative of the major archetypes of time series data. We compare the results of five different models across seven datasets. We sweep across nine groups to determine the factors that consistently affect downstream performance. 

The research questions investigated in this article: \textbf{(Q1)}~Does synthetic augmentation help when real data is
abundant? \textbf{(Q2)}~Does it help in low-resource settings? \textbf{(Q3)}~Which bundle types transfer best to real-world
forecasting? \textbf{(Q4)}~How does difficulty conditioning affect downstream performance? \textbf{(Q5)}~Do curriculum annealing and cross-channel correlation provide additional benefit?

\paragraph{Contributions.}
We contribute (1)~a controllable difficulty-conditioned synthetic time series
generator with four bundle types and Dirichlet variance allocation;
(2)~a large-scale empirical study that comprises 4,218 runs across nine experiment groups,
five architectures, and seven datasets. (3) the identification of architecture as the primary determinant
of outcome, with concrete directions for bundle selection, scheduling, and deployment.
Section~\ref{sec:method} describes the generator;
Section~\ref{sec:experiments} details the setup;
Sections~\ref{sec:results}--\ref{sec:discussion} present and interpret the findings.

\section{Related work}
\label{sec:background}
Augmenting sparse datasets with synthetic data is an established technique in
time series analysis \cite{forestier2017generating,lin2019generating,fawaz2018data}.
In the era of large foundational forecasters, synthetic data has become a central
ingredient: Amazon's Chronos2 \cite{ansari2025chronos} uses synthetic multivariate
and covariate-informed series during pre-training and even releases a version trained
exclusively on synthetic data that remains highly competitive with its real-data
counterpart; Google's TimesFM \cite{das2024decoder} adds 20\% synthetic data to its
curriculum and validates the gain via ablation; Salesforce's MOIRAI \cite{woo2024unified}
follows a similar mixed pre-training strategy.
All three achieve better or equivalent results to full-shot models in zero-shot and few-shot
scenarios.

Broader lessons from adjacent fields reinforce the case.
Synthetic data has driven gains in computer vision
\cite{paulin2023review,borkman2021unity}, healthcare
\cite{gonzales2023synthetic,giuffre2023harnessing}, and
large language models \cite{patel2024datadreamer,gan2024towards},
offering benefits such as unlimited scale, no GDPR restriction on sensitive
data \cite{liu2024preserving}, reduced data bias \cite{paproki2024synthetic},
and controlled curricula.
Research on the scaling laws governing how much synthetic data a model can
absorb has advanced rapidly for text \cite{kang2025demystifying,goyal2023synthetic},
but equivalent results for time series remain absent.

Despite this momentum, there is no consensus on how to use synthetic
data for time series: which generator types transfer, in what
quantities, and whether curriculum scheduling matters.
This paper fills that gap with a controlled, large-scale study spanning multiple
architectures and datasets.

% #############################################################################
\section{Methodology}
\label{sec:method}
% #############################################################################

The generator is organized around three components: a \emph{bundle system}
that defines the temporal character of each channel (\S\ref{subsec:bundles});
difficulty conditioning and cross-channel correlation controls that modulate
complexity and inter-variable structure (\S\ref{subsec:diff_corr}); and
data-mixing strategies that govern how synthetic and real samples are combined
during training (\S\ref{subsec:modes}).

% =============================================================================
\subsection{Bundle System}
\label{subsec:bundles}
% =============================================================================

A \emph{bundle} is a stochastic template that defines the temporal archetype
of a univariate channel: whether it is dominated by seasonality, structural
breaks, long-range dependence, or volatility clustering.
A $D$-channel series $\mathbf{X}\!\in\!\R^{T\times D}$ is constructed by
drawing a bundle type $\mathcal{B}_j\!\sim\!\mathrm{Categorical}(\mathbf{w}_{\mathcal{B}})$
independently for each channel~$j$, where $\mathbf{w}_{\mathcal{B}}$
are the mixture probabilities (uniform across the four types by default), and
sampling a univariate series from the corresponding template.

\paragraph{Variance allocation.}
Within each bundle, components are combined so that each contributes a
controlled share of total output variance.
Each component $c_k(t)$ is standardised to unit variance:
$\tilde{c}_k(t)=(c_k(t)-\bar{c}_k)/\!\sqrt{\Var(c_k)+\epsilon}$.
Mixing weights are drawn from a Dirichlet distribution:
\begin{equation}
  \mathbf{w}\sim\Dirichlet\!\bigl(\alpha(d)\cdot\boldsymbol{\tau}\bigr),
  \qquad
  \alpha(d)=(1-d)\,\alpha_{\mathrm{easy}}+d\,\alpha_{\mathrm{hard}},
\label{eq:dirichlet}
\end{equation}
with $\alpha_{\mathrm{easy}}\!=\!80$, $\alpha_{\mathrm{hard}}\!=\!6$, and
bundle-specific target fractions $\boldsymbol{\tau}$.
The output $y(t)=\sum_k\!\sqrt{w_k}\,\tilde{c}_k(t)$ has approximately
unit variance.  High concentration (low~$d$) clusters weights near
$\boldsymbol{\tau}$; low concentration (high~$d$) allows free variation.

Four bundles cover the principal real-world temporal archetypes:

\subsubsection{Seasonal-Trend (ST)}

Combines Fourier seasonality, a deterministic trend, and an autoregressive
residual:
\begin{equation}
  y_{\textsc{st}}(t)
  = \sqrt{w_s}\,\tilde{s}(t)
  + \sqrt{w_{\mathrm{tr}}}\,\widetilde{\mathrm{tr}}(t)
  + \sqrt{w_r}\,\tilde{r}(t)
  + \varepsilon(t),
\label{eq:st}
\end{equation}
where weights $w_s,w_{\mathrm{tr}},w_r$ are Dirichlet-sampled as in
\eqref{eq:dirichlet}, and $\varepsilon(t)\!\sim\!\Normal(0,\sigma_{\mathrm{obs}}^2)$
is i.i.d.\ observation noise with $\sigma_{\mathrm{obs}}$ drawn from a
difficulty-scaled range.
The seasonal component is a superposition of $K\!\in\![1,6]$ sinusoids:
\begin{equation}
  \tilde{s}(t)=\sum_{k=1}^{K}a_k\sin\!\Bigl(\tfrac{2\pi k\,t}{P}+\phi_k\Bigr),
  \quad a_k\propto k^{-\alpha},
\label{eq:fourier}
\end{equation}
with shared period $P\!\in\!\{24,48,96,168,336\}$, phases
$\phi_k\!\sim\!\mathcal{U}(0,2\pi)$, and decay exponent $\alpha$ decreasing
with difficulty~$d$ (more harmonics at higher difficulty).
The residual $\tilde{r}(t)$ is AR(1) with coefficient $\phi\!\sim\!\mathcal{U}(-0.9,0.9)$.
Trend forms are linear, quadratic (degree~2), or slow exponential
($b\!\sim\!\mathcal{U}(-0.01,0.01)$, mild by design).

\subsubsection{Non-stationary Regime (NR)}

A Markov-switching AR process with $M\!\in\!\{2,3,4\}$ regimes, where
regime means and transition probabilities are difficulty-scaled, producing
structural breaks of increasing frequency.
A stochastic trend (random walk, geometric Brownian motion, or
Ornstein--Uhlenbeck) is mixed in via Dirichlet weights~\eqref{eq:dirichlet}.

\subsubsection{Long Memory (LM)}

Slowly decaying autocorrelations via ARFIMA$(0,d_{\mathrm{frac}},0)$
($d_{\mathrm{frac}}\!\in\!(-0.45,0.45)$) or approximate fractional
Brownian motion (Hurst $H\!\in\![0.6,0.9]$), with an optional seasonal
overlay whose probability increases with~$d$.

\subsubsection{Volatility Events (VE)}

Models financial dynamics: a GARCH(1,1) process captures volatility
clustering; a Hawkes self-exciting point process adds sharp event spikes.
Baseline intensity and spike amplitude both scale with difficulty.

A representative channel from each bundle at $d=0.5$ is shown in
Appendix~\ref{app:bundle_examples}, as well as pseudo-code in Appendix\ref{app:bundle_pseudocode},.

% =============================================================================
\subsection{Difficulty and Cross-Channel Correlation}
\label{subsec:diff_corr}
% =============================================================================

\paragraph{Difficulty conditioning.}
A scalar $d\!\in\![0,1]$ governs the complexity of every generated series.
Higher $d$ yields more harmonics, stronger trends, more frequent regime
switches, and larger observation noise.  By default, $d$ is drawn from a
balanced three-component truncated-normal (TN) mixture:
\begin{equation}
  d \;\sim\;
  0.30\;\TN(0.20,0.08)
  \;+\;0.40\;\TN(0.50,0.10)
  \;+\;0.30\;\TN(0.80,0.08).
\label{eq:diff_default}
\end{equation}
For targeted ablations, four named modes fix the distribution:
\texttt{uniform} ($d\!\sim\!\mathcal{U}(0,1)$),
\texttt{easy} ($d\!\sim\!\Beta(2,5)$),
\texttt{medium} ($d\!\sim\!\Beta(2,2)$), and
\texttt{hard} ($d\!\sim\!\Beta(5,2)$).

\paragraph{Cross-channel correlation.}
Real multivariate series exhibit inter-channel correlations absent from
independently generated channels.  With probability $p_{\mathrm{latent}}$,
the generated matrix $\mathbf{X}\!\in\!\R^{T\times D}$ is transformed via
\begin{equation}
  \mathbf{Y} = \mathbf{X}\mathbf{A}^\top,
  \qquad
  \mathbf{A} = (1-\rho)\,\mathbf{I}_D + \rho\,\tilde{\mathbf{A}},
\label{eq:latent}
\end{equation}
where $\tilde{\mathbf{A}}_{ij}\!\sim\!\Normal(0,1)$ (row-normalised) and
$\rho\!\sim\!\mathcal{U}(0.20,\,0.50{+}0.20d)$ controls cross-channel
dependence strength.

% =============================================================================
\subsection{Data-Mixing Strategies}
\label{subsec:modes}
% =============================================================================

Four training modes govern the interaction between real data
$\Dcal_{\mathrm{real}}$ and synthetic data $\Dcal_{\mathrm{synth}}$.
In \textbf{real} mode (baseline), no synthetic data is used.
In \textbf{mixed} mode,
$|\Dcal_{\mathrm{synth}}|=\lfloor|\Dcal_{\mathrm{real}}^{\mathrm{orig}}|\cdot r_{\mathrm{synth}}\rfloor$
synthetic samples are concatenated with real training data, where
$|\Dcal_{\mathrm{real}}^{\mathrm{orig}}|$ is calculated based on the pre-sparsity real size
so synthetic volume is independent of data withholding.

Data scarcity is simulated by a sparsity parameter $s\!\in\!(0,1]$: a
fraction $s$ of real training windows is retained while preserving temporal
order; validation and test sets are never reduced for comparable results.

Curriculum learning is explored via two annealing modes:
\texttt{anneal} starts at $r_{\mathrm{synth}}\!=\!1$ and decays to~$0$
(synthetic$\to$real); \texttt{anneal\_inverse} does the reverse.
Both support a \emph{hard} schedule (binary switch at epoch $e_a$) and a
\emph{gradual} schedule (linear interpolation over $E$ total epochs), giving for an epoch $e$:
\begin{equation}
  r(e) =
  \begin{cases}
    r_{\mathrm{start}} & e < e_a \quad\text{(hard),} \\[3pt]
    r_{\mathrm{end}} & e \ge e_a \quad\text{(hard),} \\[3pt]
    r_{\mathrm{start}} + \dfrac{e}{E-1}(r_{\mathrm{end}}-r_{\mathrm{start}})
    & \text{(gradual).}
  \end{cases}
\label{eq:anneal}
\end{equation}
For \texttt{anneal}, $(r_{\mathrm{start}},r_{\mathrm{end}})=(1,0)$;
for \texttt{anneal\_inverse}, $(0,1)$.
In all modes, validation and test sets consist exclusively of real data.

% #############################################################################
\section{Experiments}
\label{sec:experiments}
% #############################################################################

% =============================================================================
\subsection{Datasets}
\label{subsec:datasets}
% =============================================================================

We use seven widely adopted long-term forecasting benchmarks~\cite{zhou2021informer,wu2021autoformer},
summarised in Table~\ref{tab:datasets}.
All datasets are split into train / validation / test with standard chronological partitions.

\begin{table}[h]
\centering
\caption{Real-world datasets.  $D$: number of channels; $N$: approximate
  total samples; Freq: sampling frequency.  The four datasets marked
  with~$\dagger$ are used in ablation studies (Groups~4--8); all seven
  are used in Groups~1--3.}
\label{tab:datasets}
\begin{tabular}{@{}lrrll@{}}
\toprule
Dataset & $D$ & $N$ & Freq & Domain \\
\midrule
ETTh1$^\dagger$ & 7 & 17\,420 & 1\,h & Electricity transformer \\
ETTh2           & 7 & 17\,420 & 1\,h & Electricity transformer \\
ETTm1$^\dagger$ & 7 & 69\,680 & 15\,min & Electricity transformer \\
ETTm2           & 7 & 69\,680 & 15\,min & Electricity transformer \\
Weather$^\dagger$ & 21 & 52\,696 & 10\,min & Meteorological \\
Electricity$^\dagger$ & 321 & 26\,304 & 1\,h & Power consumption \\
Traffic         & 862 & 17\,544 & 1\,h & Road occupancy \\
\bottomrule
\end{tabular}
\end{table}

% =============================================================================
\subsection{Models}
\label{subsec:models}
% =============================================================================

We evaluate five models spanning the major architectural families:
DLinear~\cite{zeng2023transformers}, SegRNN~\cite{lin2024segrnn},
TimesNet~\cite{wu2023timesnet}, iTransformer~\cite{liu2024itransformer},
and PatchTST~\cite{nie2023patchtst}. All models are trained for 10~epochs following the TSLib convention~\cite{wu2023timesnet}, with batch size~32, Adam optimizer ($\mathrm{lr}\!=\!10^{-4}$), and MSE loss. Input window $L\!=\!96$; label length~48.

\begin{table}[h]
\centering
\caption{Forecasting models and their key hyperparameters.}  
\label{tab:models}
\begin{tabular}{@{}llp{7cm}@{}}
\toprule
Model & Family & Key hyperparameters \\
\midrule
DLinear & Linear & Channel-independent (\texttt{--individual}) \\
SegRNN & RNN & $\mathrm{seg\_len}\!=\!48$, $d\!=\!512$, dropout$\!=\!0.5$ \\
TimesNet & CNN & $d\!=\!64$, $d_{\mathrm{ff}}\!=\!64$, layers$\!=\!2$, top-$k\!=\!5$ \\
iTransformer & Transformer & $d\!=\!512$, $d_{\mathrm{ff}}\!=\!512$, layers$\!=\!4$, heads$\!=\!8$ \\
PatchTST & Transformer & $d\!=\!128$, $d_{\mathrm{ff}}\!=\!256$, layers$\!=\!3$, heads$\!=\!16$, patch$\!=\!16$, stride$\!=\!8$ \\
\bottomrule
\end{tabular}
\end{table}

% =============================================================================
\subsection{Evaluation Protocol}
\label{subsec:eval}
% =============================================================================

We report MSE on the held-out real test set for prediction horizons
$H\!\in\!\{96,336\}$ (short-term and long-term), averaged over 3 independent seeds (2021--2023).
Jobs were dispatched across GPU cluster nodes using GNU Parallel~\cite{tange2011parallel}.
Group~1 trains on full real data without any alteration.
Group~2 trains on reduced real data without augmentation and quantifies the
degradation gap that synthetic data could close.
Groups~3--9 introduce synthetic augmentation under progressively controlled conditions,
detailed in Section~\ref{subsec:groups}.

% =============================================================================
\subsection{Experiment Groups}
\label{subsec:groups}
% =============================================================================

We organize $4{,}218$ runs across nine groups,
each isolating a specific factor.
Throughout, $s\!\in\!(0,1]$ denotes the fraction of real training windows
retained and $r_{\mathrm{synth}}\!\ge\!0$ the number of synthetic samples
added as a multiple of the original (pre-sparsity) dataset size.
Table~\ref{tab:groups} provides a summary.

\begin{table}[t]
\centering
\caption{Overview of the nine experiment groups.  \emph{Datasets}
indicates whether the full set (7) or the ablation subset (4) is used.
All groups use 3~seeds and 2 horizons $H\!\in\!\{96,336\}$.
Controlled parameters are listed as swept values; all others are held
at their defaults (\texttt{data\_mode}$=$\texttt{real},
$r_{\mathrm{synth}}\!=\!0$, $s\!=\!1$, difficulty$=$\texttt{uniform},
bundle$=$\texttt{all}, $p_{\mathrm{latent}}\!=\!0$).}
\label{tab:groups}
\resizebox{\textwidth}{!}{%
\begin{tabular}{@{}cl l l r@{}}
\toprule
Group & Name & Swept factor(s) & Datasets & Runs \\
\midrule
1 & Baseline         & ---                                        & Full\,(7) & 210 \\
2 & Sparsity         & $s\in\{0.10,0.25,0.50\}$                  & Full\,(7) & 630 \\
3 & Augmentation     & $r_{\mathrm{synth}}\in\{0.25,0.50,1.0\}$ (mixed) & Full\,(7) & 630 \\
4 & Low-resource     & $s\in\{0.10,0.25\}\times r_{\mathrm{synth}}\in\{0.5,1.0,2.0\}$ (mixed) & Ablation\,(4) & 720 \\
5 & Difficulty       & difficulty $\in\{\texttt{uniform},\texttt{easy},\texttt{medium},\texttt{hard}\}$ & Ablation\,(4) & 480 \\
6 & Bundle           & bundle $\in\{\textsc{st},\textsc{nr},\textsc{lm},\textsc{ve}\}$ & Ablation\,(4) & 480 \\
7 & Curriculum       & \makecell[l]{mode $\in\{\texttt{anneal},\texttt{anneal\_inverse}\}$\\$\times$ strategy $\in\{\texttt{hard},\texttt{gradual}\}$} & Ablation\,(4) & 480 \\
8 & Latent factor    & $p_{\mathrm{latent}}\in\{0.0,0.3,0.5,0.7\}$ & Ablation\,(4) & 480 \\
9 & Cache ablation   & cache\_size $\in\{0,100,500\}$ & Ablation\,(4) & 108 \\
\midrule
  & \multicolumn{3}{r}{\textbf{Total}} & \textbf{4\,218} \\
\bottomrule
\end{tabular}%
}
\end{table}

Groups~1 and 2 establish the baselines. Group~1 trains all five models on
the full real dataset to obtain reference MSE/MAE for every
(model, dataset, $H$) triple. Group~2 reduces real data to
$s\!\in\!\{0.10,0.25,0.50\}$ without augmentation, quantifying the
performance gap that synthetic data could address.

Groups~3 and 4 probe the core augmentation question. Group~3 tests
\emph{full-data augmentation} ($s\!=\!1$, $r_{\mathrm{synth}}\!\in\!\{0.25,0.5,1.0\}$,
mixed mode) across all seven datasets, asking whether adding synthetic data
helps when real data is plentiful. Group~4 is the central experiment:
real data is reduced to $s\!\in\!\{0.10,0.25\}$ while
$r_{\mathrm{synth}}\!\in\!\{0.5,1.0,2.0\}$ synthetic samples are added,
directly targeting the low-resource regime where augmentation is most
likely to help.

Groups~5--9 are single-factor ablations, each run at $r_{\mathrm{synth}}\!=\!1$,
$s\!=\!1$ on the four-dataset ablation subset.
Group~5 tests difficulty level; Group~6 isolates individual bundle types;
Group~7 evaluates curriculum annealing schedules (hard vs.\ gradual,
synth$\to$real vs.\ real$\to$synth); Group~8 sweeps the latent factor
probability $p_{\mathrm{latent}}$; Group~9 compares pre-generated sample
caches to on-the-fly generation on a subset of models (detailed results in Appendix~\ref{app:cache}).

\section{Results}
\label{sec:results}

We report MSE on held-out real test sets, averaged over three independent seeds.
Throughout this section, \emph{MSE improvement} is defined as
$(\text{MSE}_\text{base} - \text{MSE}_\text{aug})/\text{MSE}_\text{base} \times 100\%$,
so \emph{positive values mean the augmented model is better} and negative values
mean it is worse.
Baselines are stable across seeds: the median coefficient of variation across all
(model, dataset, $H$) triples is 0.40\% (median std $= 0.0013$), so the effects
reported below are not masking noisy measurements.

\paragraph{Overall picture.}
Averaged across all five architectures and all configurations in Group~3
(full-data augmentation, $s=1$), synthetic data hurts in 67\% of trials.
Augmentation is not a universal improvement.
However, the effect is strongly architecture-dependent: TimesNet and iTransformer
improve in 67\% and 71\% of trials respectively, while DLinear improves in only 7\%. The central question of this study is therefore not whether synthetic data helps, but when and by how much.

%% ============================================================
\subsection{Baseline Performance}
\label{subsec:baseline}
%% ============================================================

Full per-model, per-dataset baselines (Group~1) are reported in
Appendix~\ref{app:baseline}.
iTransformer leads on the high-channel datasets (Electricity, Traffic),
while SegRNN and PatchTST are competitive on the lower-dimensional ETT
benchmarks --- consistent with published results for these architectures.

%% ============================================================
\subsection{Full-Data Augmentation and Architecture Receptiveness (Q1)}
\label{subsec:q1}
%% ============================================================

\begin{figure}[h]
\centering
\begin{tikzpicture}
\begin{axis}[
  xbar,
  width=0.70\linewidth,
  height=5.0cm,
  bar width=10pt,
  xlabel={Mean MSE improvement over full-data baseline (\%)},
  symbolic y coords={DLinear,PatchTST,SegRNN,iTransformer,TimesNet},
  ytick=data,
  xmin=-47, xmax=14,
  enlarge y limits=0.2,
  xlabel style={font=\small},
  tick label style={font=\small},
  xtick={-35,-25,-15,-5,0,5},
  xticklabels={$-35$,$-25$,$-15$,$-5$,$0$,$+5$},
]
\addplot[xbar, fill=gray!40, draw=gray!65,
  error bars/.cd, x dir=both, x explicit,
  error bar style={draw=gray!70, line width=0.8pt}]
  coordinates {
    (-19.2,DLinear)    +- (16.0,0)
    (-8.1,PatchTST)    +- (4.5,0)
    (-6.2,SegRNN)      +- (4.3,0)
    (1.6,iTransformer) +- (3.1,0)
    (2.3,TimesNet)     +- (4.8,0)
  };
\draw[dashed, gray!60, line width=0.8pt]
  (axis cs:0,DLinear) -- (axis cs:0,TimesNet);
% Labels placed 1 unit outside each whisker to avoid overlap
\node[anchor=east, font=\scriptsize] at (axis cs:-36.5,DLinear)     {$-19.2$};
\node[anchor=east, font=\scriptsize] at (axis cs:-14.0,PatchTST)    {$-8.1$};
\node[anchor=east, font=\scriptsize] at (axis cs:-11.8,SegRNN)      {$-6.2$};
\node[anchor=west, font=\scriptsize] at (axis cs: 5.2,iTransformer) {$+1.6$};
\node[anchor=west, font=\scriptsize] at (axis cs: 8.4,TimesNet)     {$+2.3$};
\end{axis}
\end{tikzpicture}
\caption{MSE improvement from synthetic augmentation (Group~3,
  $r_{\mathrm{synth}}=1$, $s=1$).
  Positive values mean the augmented model is better than real-only training.
  Error bars show $\pm$1\,std across the 11--13 available (dataset, $H$) pairs,
  indicating consistency of the effect rather than just a mean artifact.
  TimesNet and iTransformer consistently gain; the remaining three
  architectures are consistently harmed.}
\label{fig:arch_receptiveness}
\end{figure}
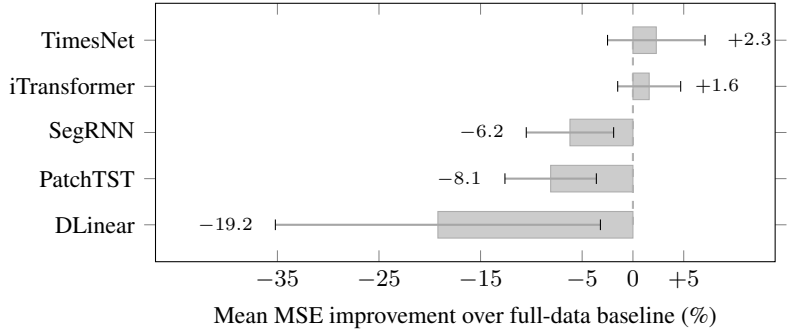

Figure~\ref{fig:arch_receptiveness} shows the mean MSE improvement
over the full-data baseline when synthetic data is added at
$r_{\mathrm{synth}}=1$ (Group~3).  Bars to the right of zero
mean the augmented model beats the real-only model; bars to the
left mean it degrades.

\textbf{Two out of five architectures consistently benefit.}
TimesNet improves in 26/39 available dataset--horizon--ratio combinations in Group~3 (67\%);
iTransformer in 27/38 available combinations (71\%).
At $r_{\mathrm{synth}}=1$ (Figure~\ref{fig:arch_receptiveness}), mean improvements
are $+2.3\%$ (TimesNet) and $+1.6\%$ (iTransformer); the remaining three degrade:
SegRNN $-6.2\%$, PatchTST $-8.1\%$, DLinear $-19.2\%$ (error bars show
$\pm$1\,std across dataset/horizon pairs).
Increasing $r_{\mathrm{synth}}$ from $0.25$ to $1.0$ provides marginal additional
gain for receptive models but deepens the penalty for non-receptive ones.

TimesNet benefits from diverse spectral characteristics found within synthetic data due to the architecture decomposing the input into 2D temporal-frequency maps. iTransformer applies attention over variate tokens, so it can leverage the inter-channel diversity in
synthetic multi-variate batches.  Channel-independent models (DLinear,
PatchTST) process each channel in isolation; synthetic data 
introduces distribution mismatch with no compensating structural benefit.

%% ============================================================
\subsection{Low-Resource Augmentation (Q2)}
\label{subsec:q2}
%% ============================================================

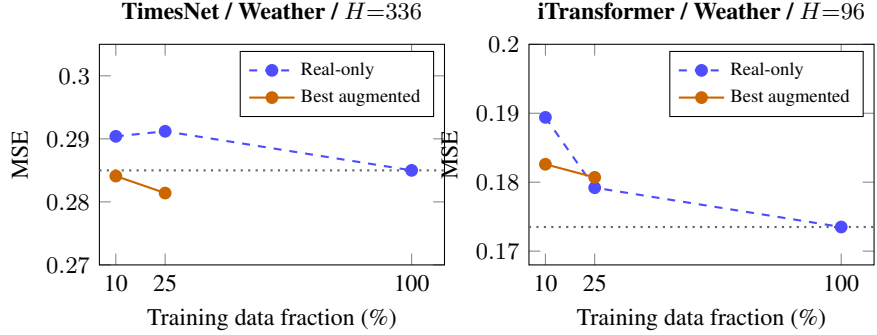
\begin{figure}[h]
\centering
\begin{tikzpicture}
\begin{axis}[
  name=leftplot,
  width=0.44\linewidth,
  height=4.5cm,
  xlabel={Training data fraction (\%)},
  ylabel={MSE},
  xtick={10,25,100},
  xticklabels={10,25,100},
  xmin=5, xmax=110,
  ymin=0.270, ymax=0.305,
  xlabel style={font=\small},
  ylabel style={font=\small},
  tick label style={font=\small},
  title style={font=\small\bfseries},
  title={TimesNet / Weather / $H{=}336$},
  legend style={font=\scriptsize, at={(0.97,0.97)}, anchor=north east},
  legend cell align=left,
]
% Real-only sparse
\addplot[mark=*, mark size=2pt, color=blue!70, thick, dashed,
  every mark/.append style={solid, fill=blue!70}]
  coordinates {(10, 0.2904) (25, 0.2912) (100, 0.2850)};
\addlegendentry{Real-only}
% Best augmented
\addplot[mark=*, mark size=2pt, color=orange!80!black, thick,
  every mark/.append style={solid, fill=orange!80!black}]
  coordinates {(10, 0.2841) (25, 0.2814)};
\addlegendentry{Best augmented}
% Full baseline
\draw[dotted, black!60, line width=0.8pt]
  (axis cs:5,0.2850) -- (axis cs:110,0.2850)
  node[right, font=\scriptsize, black!60] {full};
\end{axis}
\begin{axis}[
  name=rightplot,
  at={(leftplot.south east)},
  anchor=south west,
  xshift=0.08\linewidth,
  width=0.44\linewidth,
  height=4.5cm,
  xlabel={Training data fraction (\%)},
  ylabel={MSE},
  xtick={10,25,100},
  xticklabels={10,25,100},
  xmin=5, xmax=110,
  ymin=0.168, ymax=0.200,
  xlabel style={font=\small},
  ylabel style={font=\small},
  tick label style={font=\small},
  title style={font=\small\bfseries},
  title={iTransformer / Weather / $H{=}96$},
  legend style={font=\scriptsize, at={(0.97,0.97)}, anchor=north east},
  legend cell align=left,
]
\addplot[mark=*, mark size=2pt, color=blue!70, thick, dashed,
  every mark/.append style={solid, fill=blue!70}]
  coordinates {(10, 0.1894) (25, 0.1792) (100, 0.1735)};
\addlegendentry{Real-only}
\addplot[mark=*, mark size=2pt, color=orange!80!black, thick,
  every mark/.append style={solid, fill=orange!80!black}]
  coordinates {(10, 0.1826) (25, 0.1807)};
\addlegendentry{Best augmented}
\draw[dotted, black!60, line width=0.8pt]
  (axis cs:5,0.1735) -- (axis cs:110,0.1735)
  node[right, font=\scriptsize, black!60] {full};
\end{axis}
\end{tikzpicture}
\caption{Low-resource augmentation (Group~4).  Dashed line: real-only
  training at each sparsity level (Group~2).  Solid line: best
  synthetic augmentation ratio observed in the sweep
  ($r_{\mathrm{synth}}\in\{0.5,1,2\}$, $s\in\{0.10,0.25\}$).
  Dotted horizontal: full-data baseline (Group~1).
  Augmented points below the dotted line surpass the full-data ceiling.}
\label{fig:lowresource}
\end{figure}

Figure~\ref{fig:lowresource} shows MSE as a function of training data
fraction for the two architecturally receptive models on the Weather
dataset ($H=336$).

\textbf{Synthetic augmentation can surpass the full-data baseline.}
TimesNet trained on only 10\% of Weather with synthetic augmentation
achieves MSE~$=0.2841$, beating the full-data real-only model
($0.2850$; dashed line).  At 25\% sparsity, the augmented model
reaches $0.2814$, $1.3\%$ below the full-data baseline.
This is the only regime in our study where synthetic data provides an
unconditional win over more real data. iTransformer similarly improves MSE by $3.6\%$ relative to the sparse baseline at $s=0.10$ ($0.1826$ vs.\ sparse $0.1894$), recovering
$42.8\%$ of the sparse-to-full gap, though it does not surpass the
full-data model ($0.1735$).

Across all dataset / sparsity combinations, augmentation beats the
sparse (no-augmentation) baseline in 4/16 combinations for TimesNet
and 3/16 for iTransformer; the remaining models show fewer than 2/16
wins.  The benefits are architecture-conditional rather than
universal.

%% ============================================================
\subsection{Ablation Studies (Q3--Q5)}
\label{subsec:ablation}
%% ============================================================

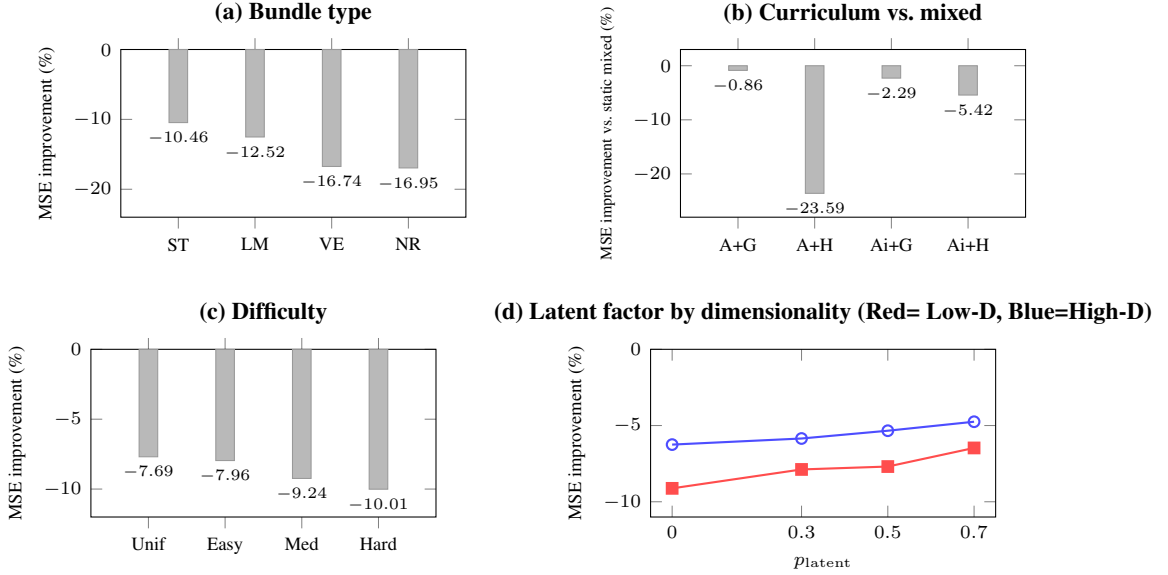
\begin{figure}[h]
\centering
\pgfplotsset{
  ablation style/.style={
    ybar,
    bar width=7pt,
    ylabel={MSE improvement (\%)},
    ymax=0,
    enlarge x limits=0.25,
    xtick=data,
    tick label style={font=\scriptsize},
    ylabel style={font=\scriptsize},
    title style={font=\small\bfseries},
    width=0.44\linewidth,
    height=3.8cm,
  }
}
% Row 1
\begin{tikzpicture}
% (a) Bundle
\begin{axis}[
  ablation style,
  symbolic x coords={ST, LM, VE, NR},
  title={(a) Bundle type},
  ymin=-24,
  nodes near coords,
  nodes near coords style={font=\tiny, rotate=0},
]
\addplot[fill=gray!55, draw=gray!80]
  coordinates {(ST,-10.46) (LM,-12.52) (VE,-16.74) (NR,-16.95)};
\end{axis}
% (b) Curriculum
\begin{axis}[
  ablation style,
  at={(0.53\linewidth,0)},
  symbolic x coords={A+G, A+H, Ai+G, Ai+H},
  title={(b) Curriculum vs.\ mixed},
  ylabel={MSE improvement vs.\ static mixed (\%)},
  ylabel style={font=\tiny},
  ymin=-28, ymax=3,
  nodes near coords,
  nodes near coords style={font=\tiny},
]
\addplot[fill=gray!55, draw=gray!80]
  coordinates {(A+G,-0.86) (Ai+G,-2.29) (Ai+H,-5.42) (A+H,-23.59)};
\end{axis}
\end{tikzpicture}

\vspace{0.3cm}

\begin{tikzpicture}
% (c) Difficulty
\begin{axis}[
  ablation style,
  symbolic x coords={Unif, Easy, Med, Hard},
  title={(c) Difficulty},
  ymin=-12, ymax=0,
  nodes near coords,
  nodes near coords style={font=\tiny},
]
\addplot[fill=gray!55, draw=gray!80]
  coordinates {(Unif,-7.69) (Easy,-7.96) (Med,-9.24) (Hard,-10.01)};
\end{axis}
% (d) Latent factor
\begin{axis}[
  at={(0.53\linewidth,0)},
  width=0.44\linewidth,
  height=3.8cm,
  xlabel={$p_{\mathrm{latent}}$},
  ylabel={MSE degradation (\%)},
  title={(d) Latent factor by dimensionality (Red= Low-D, Blue=High-D)},
  xlabel style={font=\scriptsize},
  ylabel style={font=\scriptsize},
  tick label style={font=\scriptsize},
  title style={font=\small\bfseries},
  xmin=-0.05, xmax=0.75,
  ymin=-11, ymax=0,
  ylabel={MSE improvement (\%)},
  xtick={0.0,0.3,0.5,0.7},
  legend style={font=\scriptsize, at={(0.03,0.03)}, anchor=south west,
    legend columns=2, column sep=4pt},
  legend cell align=left,
]
\addplot[mark=o, thick, color=blue!70]
  coordinates {(0.0,-6.25) (0.3,-5.85) (0.5,-5.34) (0.7,-4.75)};
\addplot[mark=square*, thick, color=red!70]
  coordinates {(0.0,-9.12) (0.3,-7.88) (0.5,-7.69) (0.7,-6.47)};
\end{axis}
\end{tikzpicture}

\caption{Ablation results (Groups~5--8, $s=1$, $r=1$, averaged over all architectures).
\emph{All y-axes show MSE improvement vs.\ the full-data baseline}: lower (more negative) values indicate greater degradation.
(a)~Every bundle type degrades MSE on average; ST causes the least harm.
(b)~Curriculum strategies vs.\ static mixed at $r=1$: gradual annealing
performs the same as static mixing; the hard switch (\textbf{A+H}) is catastrophic.
A\,=\,\texttt{anneal}, Ai\,=\,\texttt{anneal\_inverse}, G\,=\,gradual, H\,=\,hard.
(c)~Difficulty mode: negligible spread ($<2.4$ pp); uniform is marginally best.
(d)~Latent factor: higher $p_{\mathrm{latent}}$ reduces the degradation for
both high-D datasets (Weather 21-ch, Electricity 321-ch) and low-D ETT datasets.}
\label{fig:ablation}
\end{figure}

Figure~\ref{fig:ablation} summarises the four ablation groups.
All bars show \emph{MSE improvement} relative to the full-data baseline
(Groups~5--8, $s=1$, $r_{\mathrm{synth}}=1$): lower (more negative) bars indicate greater degradation.
The ablations use a fixed mixed-mode setting and test one factor at a time;
because architecture choice is held fixed across all models, the averages
include both receptive and non-receptive architectures, which is why all
conditions show net degradation.  The comparisons are \emph{relative}: the
question here is which bundle, schedule, or difficulty minimises the harm.

\paragraph{Bundle type (Q3).}
No bundle type produces a net improvement on average.  The least harmful
choice is Seasonal-Trend (ST), with a mean degradation of $10.5\%$, compared
to $12.5\%$ for LM, $16.7\%$ for VE, and $17.0\%$ for NR.
The ranking is consistent across datasets: ST $<$ LM $<$ VE $\approx$ NR.
NR and VE bundles incorporate abrupt regime switches and volatility clusters
that are structurally dissimilar from the smooth, trend-dominated patterns in
ETT and Weather, which causes the largest distribution mismatch.

\paragraph{Curriculum scheduling (Q5).}
Results here are relative to static mixed augmentation ($r_{\mathrm{synth}}=1$,
Group~3) rather than the full-data baseline, to isolate the effect of scheduling.
Gradual synth$\to$real annealing (\texttt{anneal+gradual}) performs essentially
the same as static mixing ($+0.9\%$), meaning curriculum ordering
does not help but does no additional harm when applied gradually.
The hard-switch variant (\texttt{anneal+hard}) is catastrophic: MSE increases
$23.6\%$ above static mixing due to the abrupt distribution shift at epoch~5
mid-training.

\paragraph{Difficulty conditioning (Q4).}
All four difficulty modes degrade MSE, clustering between $7.7\%$ and $10.0\%$
above baseline.  The differences are small ($<2.4$ pp), suggesting difficulty
is a minor factor.  Uniform difficulty performs marginally best; harder
distributions incur slightly larger penalties, consistent with noisier gradients.

\paragraph{Latent factor (Q5).}
The latent factor wrapper reduces the degradation monotonically for
high-dimensional datasets (Weather: 21 channels; Electricity: 321 channels):
from $6.3\%$ at $p_{\mathrm{latent}}=0$ to $4.8\%$ at $p=0.7$.
For low-dimensional ETT datasets (7 channels), the reduction is larger in
absolute percentage points ($9.1\%$ to $6.5\%$), although the high-dimensional
datasets remain less degraded overall.  Higher $p_{\mathrm{latent}}$ is beneficial at no cost;
$p=0.5$ is a safe default.

%% ============================================================
\subsection{Cache Efficiency (Group~9)}
\label{subsec:cache}
%% ============================================================

A practical concern when deploying synthetic augmentation is the cost of
on-the-fly generation.  We compare on-the-fly generation (cache size~0)
against fixed pre-generated caches of 100 and 500 samples.
Across 12 fully observed (model, dataset, $H$) combinations, the maximum
MSE difference between any two cache strategies is $0.022$, with a
median of $0.009$.  A fixed cache of 500 samples is practically
comparable to on-the-fly generation and reduces per-epoch
generation overhead to a one-time cost. Detailed results on the cache ablation can be found in Appendix \ref{app:cache}.

%% ============================================================
\section{Discussion}
\label{sec:discussion}
%% ============================================================

\paragraph{Synthetic data in LLMs versus time series fine-tuning.}
In large language models, synthetic data has proved most valuable in two roles:
pre-training at scale (filling gaps in rare phenomena, extending coverage
of low-resource languages or domains) and alignment (generating trajectories for RL).
In both cases the model is large, is exposed to many domains simultaneously, and
the synthetic distribution is only a small fraction of the total training signal.
The closest time-series analogue is the pre-training regime used by Chronos2,
TimesFM, and MOIRAI: a large transformer is trained from
scratch on millions of series drawn from both real and synthetic sources,
before zero-shot or few-shot transfer to downstream tasks.
Our study tests a different but complementary setting: synthetic augmentation
within a dataset full-shot of a model that starts from random weights.
Here the model has far fewer epochs to reconcile a synthetic distribution with the
target domain, and the regularisation effect that makes augmentation work in
vision or language is counteracted by local distribution mismatch.

\paragraph{Reconciling with Chronos2.}
Amazon's KernelSynth ablation reports that a version of Chronos2 trained only on synthetic data remains competitive with the real-data model.
Three structural differences explain this: (i)~Chronos2 is a channel-joint decoder-only transformer, the receptive architecture our results flag as beneficial; (ii)~KernelSynth is used in pre-training, not per-dataset fine-tuning, so the model never overfits to a single distribution; (iii)~millions of synthetic series versus hundreds in our runs smooth out generator artifacts at scale.
These conditions together place Chronos2 squarely in the regime where our study predicts synthetic data to help.

\paragraph{Why do three of four bundles fail?}
The NR, LM, and VE bundles consistently degrade performance more than ST across
all tested datasets.
We believe this is a distribution mismatch problem rather than an
implementation defect.
The benchmark datasets used here are dominated
by seasonal-trend dynamics; NR introduces abrupt regime switches,
VE introduces GARCH-style volatility clusters, and LM introduces long-memory
autocorrelation, none of which are prominent in either ETTh, Electricity or Weather.
From a regularisation perspective, these bundles add irrelevant structure: the
model must allocate capacity to patterns it ends up not encountering in the test set.
ST succeeds because it directly reinforces the inductive bias of the target domain.
Bundle selection should therefore be conditioned on the statistical profile of the target dataset: a financial dataset with prominent regime changes and volatility clusters could reverse the ranking.

\paragraph{Architecture as the key gating factor.}
A consistent theme across Groups 3-8 is that architecture dominates all other
factors.  Even with the best bundle (ST), best schedule (gradual), and best
latent factor ($p=0.7$), DLinear and PatchTST do not benefit.
Channel-independent models process each variate in isolation, so they cannot
leverage the cross-variate diversity that synthetic multi-variate batches provide.
This observation points directly toward a new generation of targets:
foundational architectures that are designed to learn from many channels
jointly, such as multi-variate pre-trained transformers.
These models operate in the pre-training regime, have the capacity to absorb
distributional variation, and are channel-aware by design, matching the
conditions under which our results suggest synthetic data helps.

\paragraph{Limitations.}
This study covers five architectures and seven long-term forecasting benchmarks
with stable, trend-dominated dynamics (ETT, Weather, Electricity, Traffic).
Datasets with pronounced regime shifts, heavy tails, or sparse observations, for instance found in Finance or Healthcare are absent; the usefulness of each bundle may differ there.
All models are trained from random weights for 10 epochs, matching the TSLib
convention; foundation-model pre-training or few-shot fine-tuning may respond
differently to synthetic augmentation.
Results are evaluated on point-forecasting metrics (MSE and MAE);
probabilistic forecasting, classification, anomaly-detection, and
out-of-distribution stress tests are outside this study's scope.

\paragraph{Towards improved augmentation.}
Hard-switch curriculum schedules should be avoided entirely.
The negative results identify three correctable failure modes:
(i)~automatic spectral profiling to select the best-aligned bundle;
(ii)~moment-matching calibration of generator parameters to the target domain;
(iii)~multi-dataset pre-training to reduce sensitivity to individual bundle choice.

%% ============================================================
\section{Conclusion}
\label{sec:conclusion}
%% ============================================================

We presented a large-scale empirical study of synthetic time series augmentation
across 4,218 runs and nine experiment groups.
The central finding is that the effect of synthetic data
is architecture-conditional: channel-mixing models (TimesNet,
iTransformer) benefit consistently across trials, while channel-independent
models (DLinear, PatchTST) are reliably harmed.
The 67\% aggregate hurt rate reflects the prevalence of channel-independent
architectures in the benchmark; within the receptive class, augmentation works.
Three actionable guidelines emerge:

\begin{enumerate}[label=(\arabic*)]
\item \textbf{Prioritise receptive architectures.}  CNN- and
  multi-variate-attention-based models (TimesNet, iTransformer) benefit
  from synthetic augmentation across the board; channel-independent
  models (DLinear, PatchTST) are consistently harmed.
  Architecture selection is the strongest single predictor of outcome.

\item \textbf{Use ST bundle, gradual schedule.}  The Seasonal-Trend
  bundle is the safest choice across the tested long-term forecasting
  benchmarks.  Curriculum
  annealing should use a gradual synth$\to$real schedule; hard switches
  cause catastrophic degradation ($+24\%$ over static mixing).

\item \textbf{Synthetic data can help in low-resource settings.}  With
  ${\le}25\%$ real training data, a receptive architecture can match or
  surpass full-data performance in some (sparsity, dataset)
  combinations. A fixed cache of 500 pre-generated samples is sufficient in our cache study and adds no
  inference-time cost.
\end{enumerate}

\medskip\noindent\textbf{Acknowledgements.}
The results presented in this article were obtained on the Jülich Supercomputing Centre \cite{jsc}.
The authors thank the JSC team for their support.

\bibliographystyle{plainnat}
\bibliography{references}

@String { ACM      = {The OX Association for Computing Machinery} }

@article{r,
author = {Grand View Research},
journal = {Available online: http://www.gsi-alliance.org/wpcontent/uploads/},
pages = {3},
title = {[14] GSIA. Global Sustainable Investment Alliance Investment Review. 2018},
volume = {2018},
year={2018},
url = {https://www.grandviewresearch.com/industry-analysis/assetmanagement-market},
unidentified = {2019/03/GSIR_Review.28.pdf},
}

@unpublished{i,
author = {Idzelis, Christine},
title = {AI-Powered Hedge Funds Vastly Outperformed, Research Shows},
year={2020},
url = {https://www.institutionalinvestor.com/article/b1mssrswn1mpr0/AI-Powered-HedgeFunds-Vastly-Outperformed-Research-Shows},
}

@article{wu2021autoformer,
  title={Autoformer: Decomposition transformers with auto-correlation for long-term series forecasting},
  author={Wu, Haixu and Xu, Jiehui and Wang, Jianmin and Long, Mingsheng},
  journal={Advances in neural information processing systems},
  volume={34},
  pages={22419--22430},
  year={2021}
}

@inproceedings{zhou2021informer,
  title={Informer: Beyond efficient transformer for long sequence time-series forecasting},
  author={Zhou, Haoyi and Zhang, Shanghang and Peng, Jieqi and Zhang, Shuai and Li, Jianxin and Xiong, Hui and Zhang, Wancai},
  booktitle={Proceedings of the AAAI conference on artificial intelligence},
  volume={35},
  number={12},
  pages={11106--11115},
  year={2021}
}

@inproceedings{wu2023timesnet,
  title={TimesNet: Temporal 2D-Variation Modeling for General Time Series Analysis},
  author={Haixu Wu and Tengge Hu and Yong Liu and Hang Zhou and Jianmin Wang and Mingsheng Long},
  booktitle={International Conference on Learning Representations},
  year={2023},
}

@misc{itransformer,
  title = {stable-baseline-3},
  author = {Liu, Yong and Hu, Tengge and Zhang, Haoran and Wu, Haixu and Wang, Shiyu and Ma, Lintao and Long, Mingsheng},
  year={2024},
  howpublished = {\\\url{https://github.com/thuml/Time-Series-Library/blob/main/models/iTransformer.py}},
}

@inproceedings{zeng2023transformers,
  title={Are transformers effective for time series forecasting?},
  author={Zeng, Ailing and Chen, Muxi and Zhang, Lei and Xu, Qiang},
  booktitle={Proceedings of the AAAI conference on artificial intelligence},
  volume={37},
  number={9},
  pages={11121--11128},
  year={2023}
}

@inproceedings{kang2025demystifying,
  title={Demystifying synthetic data in llm pre-training: A systematic study of scaling laws, benefits, and pitfalls},
  author={Kang, Feiyang and Ardalani, Newsha and Kuchnik, Michael and Emad, Youssef and Elhoushi, Mostafa and Sengupta, Shubhabrata and Li, Shang-Wen and Raghavendra, Ramya and Jia, Ruoxi and Wu, Carole-Jean},
  booktitle={Proceedings of the 2025 Conference on Empirical Methods in Natural Language Processing},
  pages={10750--10769},
  year={2025}
}

@article{paulin2023review,
  title={Review and analysis of synthetic dataset generation methods and techniques for application in computer vision},
  author={Paulin, Goran and Ivasic-Kos, Marina},
  journal={Artificial intelligence review},
  volume={56},
  number={9},
  pages={9221--9265},
  year={2023},
  publisher={Springer}
}

@article{borkman2021unity,
  title={Unity perception: generate synthetic data for computer vision},
  author={Borkman, Steve and Crespi, Adam and Dhakad, Saurav and Ganguly, Sujoy and Hogins, Jonathan and Jhang, You-Cyuan and Kamalzadeh, Mohsen and Li, Bowen and Leal, Steven and Parisi, Pete and others},
  journal={arXiv preprint arXiv:2107.04259},
  year={2021}
}

@article{paproki2024synthetic,
  title={Synthetic data for deep learning in computer vision \& medical imaging: A means to reduce data bias},
  author={Paproki, Anthony and Salvado, Olivier and Fookes, Clinton},
  journal={ACM Computing Surveys},
  volume={56},
  number={11},
  pages={1--37},
  year={2024},
  publisher={ACM New York, NY}
}

@article{gonzales2023synthetic,
  title={Synthetic data in health care: A narrative review},
  author={Gonzales, Aldren and Guruswamy, Guruprabha and Smith, Scott R},
  journal={PLOS Digital Health},
  volume={2},
  number={1},
  pages={e0000082},
  year={2023},
  publisher={Public Library of Science San Francisco, CA USA}
}

@article{giuffre2023harnessing,
  title={Harnessing the power of synthetic data in healthcare: innovation, application, and privacy},
  author={Giuffr{\`e}, Mauro and Shung, Dennis L},
  journal={NPJ digital medicine},
  volume={6},
  number={1},
  pages={186},
  year={2023},
  publisher={Nature Publishing Group UK London}
}

@article{patel2024datadreamer,
  title={Datadreamer: A tool for synthetic data generation and reproducible llm workflows},
  author={Patel, Ajay and Raffel, Colin and Callison-Burch, Chris},
  journal={arXiv preprint arXiv:2402.10379},
  year={2024}
}

@article{gan2024towards,
  title={Towards a theoretical understanding of synthetic data in llm post-training: A reverse-bottleneck perspective},
  author={Gan, Zeyu and Liu, Yong},
  journal={arXiv preprint arXiv:2410.01720},
  year={2024}
}

@article{liu2024preserving,
  title={Preserving privacy in healthcare: A systematic review of deep learning approaches for synthetic data generation},
  author={Liu, Yintong and Acharya, U Rajendra and Tan, Jen Hong},
  journal={Computer Methods and Programs in Biomedicine},
  pages={108571},
  year={2024},
  publisher={Elsevier}
}

@article{goyal2023synthetic,
  title={Synthetic data revolutionizes rare disease research: How large language models and generative AI are overcoming data scarcity and privacy challenges},
  author={Goyal, Mahesh Kumar and Chaturvedi, Rahul},
  journal={International Journal on Recent and Innovation Trends in Computing and Communication},
  volume={11},
  number={11},
  pages={1368--1380},
  year={2023}
}

@article{ansari2025chronos,
  title={Chronos-2: From univariate to universal forecasting},
  author={Ansari, Abdul Fatir and Shchur, Oleksandr and K{\"u}ken, Jaris and Auer, Andreas and Han, Boran and Mercado, Pedro and Rangapuram, Syama Sundar and Shen, Huibin and Stella, Lorenzo and Zhang, Xiyuan and others},
  journal={arXiv preprint arXiv:2510.15821},
  year={2025}
}

@inproceedings{das2024decoder,
  title={A decoder-only foundation model for time-series forecasting},
  author={Das, Abhimanyu and Kong, Weihao and Sen, Rajat and Zhou, Yichen},
  booktitle={Forty-first International Conference on Machine Learning},
  year={2024}
}

@article{woo2024unified,
  title={Unified training of universal time series forecasting transformers},
  author={Woo, Gerald and Liu, Chenghao and Kumar, Akshat and Xiong, Caiming and Savarese, Silvio and Sahoo, Doyen},
  year={2024},
  publisher={PMLR}
}

@inproceedings{forestier2017generating,
  title={Generating synthetic time series to augment sparse datasets},
  author={Forestier, Germain and Petitjean, Fran{\c{c}}ois and Dau, Hoang Anh and Webb, Geoffrey I and Keogh, Eamonn},
  booktitle={2017 IEEE international conference on data mining (ICDM)},
  pages={865--870},
  year={2017},
  organization={IEEE}
}

@article{lin2019generating,
  title={Generating high-fidelity, synthetic time series datasets with doppelganger},
  author={Lin, Zinan and Jain, Alankar and Wang, Chen and Fanti, Giulia and Sekar, Vyas},
  journal={arXiv preprint arXiv:1909.13403},
  year={2019}
}

@article{fawaz2018data,
  title={Data augmentation using synthetic data for time series classification with deep residual networks},
  author={Fawaz, Hassan Ismail and Forestier, Germain and Weber, Jonathan and Idoumghar, Lhassane and Muller, Pierre-Alain},
  journal={arXiv preprint arXiv:1808.02455},
  year={2018}
}

@article{lin2024segrnn,
  title={{SegRNN}: Segment Recurrent Neural Network for Long-Term Time Series Forecasting},
  author={Lin, Shengsheng and Lin, Weiwei and Wu, Wentai and Zhao, Feiyu and Mo, Ruichao and Zhang, Haotong},
  journal={arXiv preprint arXiv:2308.11200},
  year={2024}
}

@inproceedings{liu2024itransformer,
  title={i{T}ransformer: Inverted Transformers Are Effective for Time Series Forecasting},
  author={Liu, Yong and Hu, Tengge and Zhang, Haoran and Wu, Ling and Wang, Shengnan and Luo, Lintao and Long, Mingsheng},
  booktitle={International Conference on Learning Representations},
  year={2024}
}

@inproceedings{nie2023patchtst,
  title={A Time Series is Worth 64 Words: Long-term Forecasting with Transformers},
  author={Nie, Yuqi and H. Nguyen, Nam and Sinthong, Phanwadee and Kalagnanam, Jayant},
  booktitle={International Conference on Learning Representations},
  year={2023}
}

@article{tange2011parallel,
  title={{GNU Parallel}: The Command-Line Power Tool},
  author={Tange, Ole},
  journal={;login: The USENIX Magazine},
  volume={36},
  number={1},
  pages={42--47},
  year={2011}
}

@misc{jsc,                                                          
    author       = {{Jülich Supercomputing Centre}},
    title        = {{JUWELS Cluster and Booster: Exascale pathfinder with modular supercomputing         
  architecture at Jülich Supercomputing Centre}},                                                        
    year         = {2021},                                                                               
    howpublished = {Journal of large-scale research facilities, 7(A138)},                                
    url          = {https://doi.org/10.17815/jlsrf-7-183}                                                
  }

\newpage
%% ============================================================
\appendix
%% ============================================================

%% ============================================================
\section{Baseline MSE (Group~1)}
\label{app:baseline}

Table~\ref{tab:baseline} reports full-data baseline MSE (mean over 3~seeds)
for all five models, seven datasets, and both prediction horizons.
Bold marks the best (lowest MSE) per (dataset,~$H$) pair.

\begin{table}[h]
\centering
\caption{Baseline MSE (Group~1, real-only, full data).  Mean over 3 seeds.
  Bold: best (lowest MSE) per (dataset, $H$) pair.}
\label{tab:baseline}
\setlength{\tabcolsep}{4pt}
\begin{tabular}{@{}lc rrrrr@{}}
\toprule
Dataset & $H$ & DLinear & SegRNN & TimesNet & iTransformer & PatchTST \\
\midrule
\multirow{2}{*}{ETTh1}
  & 96  & 0.396 & \textbf{0.374} & 0.402 & 0.396 & 0.390 \\
  & 336 & 0.487 & \textbf{0.455} & 0.531 & 0.489 & 0.468 \\
\multirow{2}{*}{ETTh2}
  & 96  & 0.345 & 0.293 & 0.332 & 0.312 & \textbf{0.287} \\
  & 336 & 0.592 & 0.424 & 0.480 & 0.441 & \textbf{0.418} \\
\multirow{2}{*}{ETTm1}
  & 96  & 0.346 & \textbf{0.331} & 0.339 & 0.350 & 0.337 \\
  & 336 & 0.414 & \textbf{0.398} & 0.432 & 0.461 & 0.409 \\
\multirow{2}{*}{ETTm2}
  & 96  & 0.193 & \textbf{0.173} & 0.187 & 0.185 & 0.177 \\
  & 336 & 0.382 & \textbf{0.295} & 0.319 & 0.320 & 0.304 \\
\multirow{2}{*}{Weather}
  & 96  & 0.196 & \textbf{0.165} & 0.171 & 0.174 & 0.182 \\
  & 336 & 0.282 & \textbf{0.271} & 0.285 & 0.280 & 0.282 \\
\multirow{2}{*}{Electricity}
  & 96  & 0.210 & 0.193 & 0.204 & \textbf{0.161} & 0.212 \\
  & 336 & 0.223 & 0.224 & 0.263 & \textbf{0.191} & 0.233 \\
\multirow{2}{*}{Traffic}
  & 96  & 0.696 & 0.781 & 0.640 & \textbf{0.440} & 0.555 \\
  & 336 & 0.653 & 0.818 & 0.705 & \textbf{0.479} & 0.562 \\
\bottomrule
\end{tabular}
\end{table}

%% ============================================================
\section{Bundle Examples}
\label{app:bundle_examples}

\begin{figure}[H]
\centering
\includegraphics[width=0.86\linewidth]{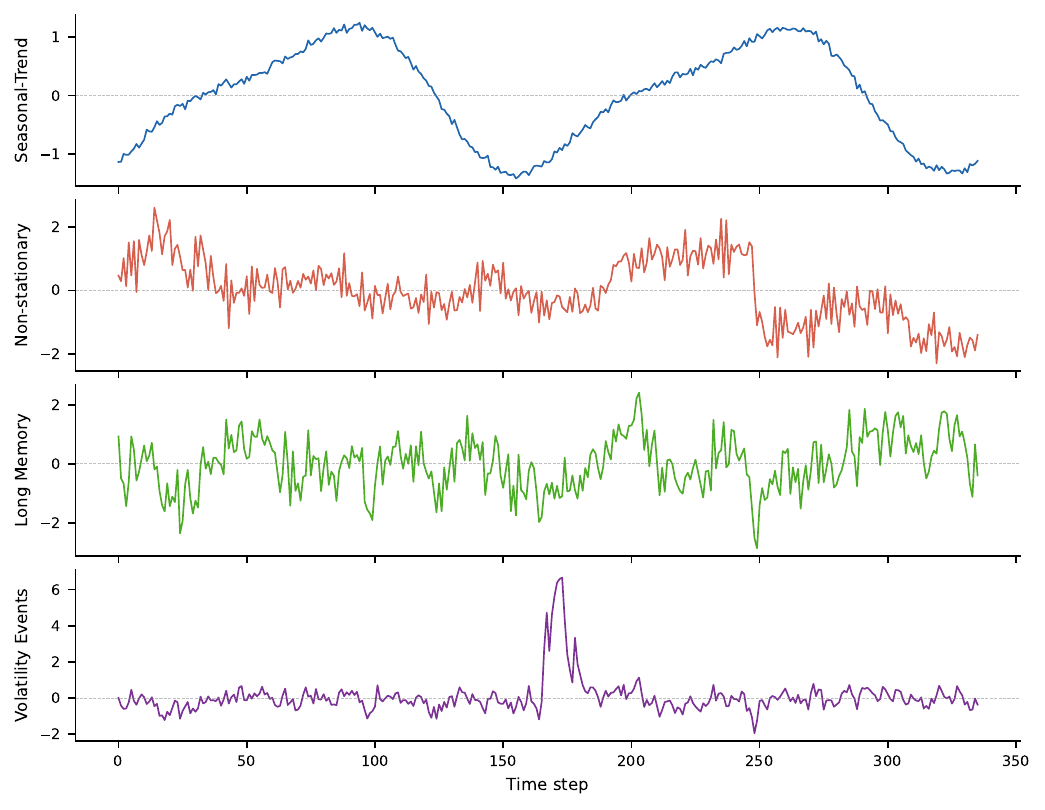}
\caption{One representative channel drawn from each bundle type at medium
  difficulty ($d=0.5$, seed fixed for reproducibility, $T=336$ time steps).
  \textbf{Seasonal-Trend (ST):} dominant periodic structure with mild AR
  residual.
  \textbf{Non-stationary (NR):} Markov-switching regime changes with a
  stochastic trend component.
  \textbf{Long Memory (LM):} slowly decaying autocorrelations producing
  persistent, non-periodic undulations.
  \textbf{Volatility Events (VE):} low-amplitude baseline punctuated by a
  sharp GARCH/Hawkes spike.}
\label{fig:bundle_examples}
\end{figure}

\section{Bundle Pseudo-code}
\label{app:bundle_pseudocode}

\medskip
\noindent 
$\tilde{x}$ denotes $x$ normalised to unit variance.
Dirichlet weights $(w_i)\sim\mathrm{Dir}(\alpha(d))$ use difficulty-scaled
targets and concentration; harder difficulties flatten the mixture and
raise the noise floor $\sigma_\mathrm{obs}(d)$.

\begin{algorithm}[H]
\caption{Seasonal-Trend (ST) bundle}
\begin{algorithmic}[1]
\Require length $T$, difficulty $d\in[0,1]$
\State $K \sim \mathrm{Discrete}\{1,\ldots,6\}$;\quad $P \sim \{24,48,96,168,336\}$;\quad $\alpha \leftarrow \alpha(d)$ \Comment{$\alpha\!\downarrow$ as $d\!\uparrow$}
\For{$k = 1,\ldots,K$}
  \State $\phi_k \sim \mathcal{U}(0,2\pi)$;\quad $a_k \leftarrow k^{-\alpha}$ \Comment{phase and power-law amplitude}
\EndFor
\State $s(t) \leftarrow \sum_{k=1}^{K} a_k \sin(2\pi k t / P + \phi_k)$ \Comment{seasonal component}
\State $\phi \sim \mathcal{U}(-0.9,\,0.9)$;\quad generate $r(t)$ as $\mathrm{AR}(1)$ with coefficient $\phi$ \Comment{residual}
\State type $\sim \mathrm{Uniform}\{\mathrm{linear},\;\mathrm{quadratic},\;\mathrm{exponential}\}$
\If{type $=$ exponential} sample $b \sim \mathcal{U}(-0.01, 0.01)$ \EndIf
\State compute $\mathrm{tr}(t)$ according to sampled type \Comment{trend}
\State $(w_s,w_{tr},w_r) \sim \mathrm{Dir}(\alpha(d)\boldsymbol{\tau}_\mathrm{ST}(d))$ \Comment{targets $[0.92,0.06,0.02]\!\to\![0.50,0.20,0.30]$ easy$\to$hard}
\State $\tilde{y}(t) \leftarrow \sqrt{w_s}\,\tilde{s}(t) + \sqrt{w_{tr}}\,\widetilde{\mathrm{tr}}(t) + \sqrt{w_r}\,\tilde{r}(t)$ \Comment{unit-variance mix}
\State $\sigma_\mathrm{obs} \leftarrow \sigma_\mathrm{obs}(d)$;\quad $y(t) \leftarrow \tilde{y}(t) + \varepsilon(t)$,\; $\varepsilon(t)\!\sim\!\mathcal{N}(0,\sigma_\mathrm{obs}^2)$
\State \Return $y$
\end{algorithmic}
\end{algorithm}

\begin{algorithm}[H]
\caption{Non-stationary Regime (NR) bundle}
\begin{algorithmic}[1]
\Require $T$, $d$
\State $M \sim \mathrm{Uniform}\{2,3,4\}$ \Comment{number of regimes}
\State $\mu_m \sim \mathcal{N}(0,\sigma_\mu(d))$,\; $\beta_m \sim \mathcal{N}(0,\sigma_\beta(d))$ for $m=1,\ldots,M$ \Comment{regime params}
\State $p_\mathrm{stay} \leftarrow p_\mathrm{stay}(d)$ \Comment{$p_\mathrm{stay}\!\downarrow$ as $d\!\uparrow$, giving more frequent transitions}
\State simulate Markov chain: $s_1,\ldots,s_T$ with $\Pr(s_{t+1}=m\mid s_t=m)=p_\mathrm{stay}$
\State $g(t) \leftarrow \mathrm{AR}(1)$ in regime $s_t$ with mean $\mu_{s_t}$ and slope $\beta_{s_t}$ \Comment{regime process}
\State type $\sim \mathrm{Uniform}\{\mathrm{random\ walk},\;\mathrm{GBM},\;\mathrm{Ornstein{-}Uhlenbeck}\}$;\; compute $\mathrm{st}(t)$
\State generate $\mathrm{AR}(1)$ residual $r(t)$
\State $(w_g,w_{st},w_r) \sim \mathrm{Dir}(\alpha(d)\boldsymbol{\tau}_\mathrm{NR}(d))$
\State $\varepsilon(t)\!\sim\!\mathcal{N}(0,\sigma_\mathrm{obs}^2(d))$
\State $y(t) \leftarrow \sqrt{w_g}\,\tilde{g}(t) + \sqrt{w_{st}}\,\widetilde{\mathrm{st}}(t) + \sqrt{w_r}\,\tilde{r}(t) + \varepsilon(t)$
\State \Return $y$
\end{algorithmic}
\end{algorithm}

\begin{algorithm}[H]
\caption{Long Memory (LM) bundle}
\begin{algorithmic}[1]
\Require $T$, $d$
\State $d_\mathrm{frac} \sim \mathcal{U}(-0.45,\,0.45)$ \textbf{or} $H \sim \mathcal{U}(0.6,\,0.9)$ for the fBm variant
\State $\mathrm{lm}(t) \leftarrow \mathrm{ARFIMA}(0,\,d_\mathrm{frac},\,0)$ \Comment{slowly decaying autocorrelations}
\State $p_\mathrm{seas}(d) \leftarrow 5\%\;\text{(easy)},\;25\%\;\text{(medium)},\;45\%\;\text{(hard)}$
\If{$\mathcal{U}(0,1) < p_\mathrm{seas}(d)$}
  \State generate Fourier seasonal overlay $s(t)$ at reduced difficulty $0.3d$
\Else
  \State $s(t) \leftarrow 0$
\EndIf
\State $n(t) \sim \mathcal{N}(0, \sigma_n^2(d))$ \Comment{noise component}
\State $(w_\mathrm{lm}, w_s, w_n) \sim \mathrm{Dir}(\alpha(d)\boldsymbol{\tau}_\mathrm{LM}(d))$
\State $y(t) \leftarrow \sqrt{w_\mathrm{lm}}\,\widetilde{\mathrm{lm}}(t) + \sqrt{w_s}\,\tilde{s}(t) + \sqrt{w_n}\,\tilde{n}(t)$
\State \Return $y$
\end{algorithmic}
\end{algorithm}

\begin{algorithm}[H]
\caption{Volatility Events (VE) bundle}
\begin{algorithmic}[1]
\Require $T$, $d$
\State sample GARCH$(1,1)$ params $(\omega,\alpha_G,\beta_G)$ scaled by $d$ \Comment{$\omega,\alpha_G\!\uparrow$ with $d$}
\State initialise $g(0)$ and $\sigma^2(0)$ from the stationary variance
\For{$t = 1,\ldots,T$}
  \State $\sigma^2(t) \leftarrow \omega + \alpha_G\,g^2(t{-}1) + \beta_G\,\sigma^2(t{-}1)$
  \State $g(t) \sim \mathcal{N}(0,\,\sigma^2(t))$ \Comment{GARCH innovation}
\EndFor
\State $\lambda_0 \leftarrow \lambda_0(d)$,\; $A \leftarrow A(d)$,\; $\delta \leftarrow \delta(d)$ \Comment{Hawkes params, all $\uparrow$ with $d$}
\State $h(t) \leftarrow \lambda_0 + A\sum_{\tau < t}\exp(-\delta(t-\tau))$ \Comment{self-exciting intensity}
\State sample event times $\{t_\mathrm{ev}\}$ from inhomogeneous $\mathrm{Poisson}(h(t))$
\State $\mathrm{spikes}(t) \leftarrow A_\mathrm{spike}(d)\sum_\mathrm{events} \kappa(t - t_\mathrm{ev})$ \Comment{$\kappa$: exponential decay kernel}
\State $n(t)\sim\mathcal{N}(0,\sigma_\mathrm{obs}^2(d))$;\quad $(w_g,w_\mathrm{sp},w_n)\sim\mathrm{Dir}(\alpha(d)\boldsymbol{\tau}_\mathrm{VE}(d))$
\State $y(t) \leftarrow \sqrt{w_g}\,\tilde{g}(t)+\sqrt{w_\mathrm{sp}}\,\widetilde{\mathrm{spikes}}(t)+\sqrt{w_n}\,\tilde{n}(t)$
\State \Return $y$
\end{algorithmic}
\end{algorithm}

\section{Cache Ablation: Detailed Results (Group~9)}
\label{app:cache}

Table~\ref{tab:cache} reports MSE (mean over 3~seeds) for each fully
observed (model, dataset, $H$) combination in Group~9, along with the
maximum pairwise difference across the three cache strategies.
All twelve combinations show a maximum difference below $0.03$, confirming
that a fixed 500-sample cache is a practical drop-in replacement for
on-the-fly generation.

\begin{table}[H]
\centering
\caption{Group~9 cache ablation.  MSE averaged over 3 seeds.
  Max~$\Delta$: largest pairwise MSE difference across the three cache sizes.}
\label{tab:cache}
\setlength{\tabcolsep}{5pt}
\begin{tabular}{@{}llc rrr r@{}}
\toprule
Model & Dataset & $H$ & Cache-0 & Cache-100 & Cache-500 & Max~$\Delta$ \\
\midrule
PatchTST       & ETTh1   &  96 & 0.4674 & 0.4670 & 0.4507 & 0.0167 \\
PatchTST       & ETTh1   & 336 & 0.5559 & 0.5339 & 0.5343 & 0.0220 \\
PatchTST       & ETTm1   &  96 & 0.3894 & 0.3913 & 0.3975 & 0.0081 \\
PatchTST       & ETTm1   & 336 & 0.4904 & 0.4795 & 0.4822 & 0.0109 \\
PatchTST       & Weather &  96 & 0.1920 & 0.1961 & 0.1925 & 0.0041 \\
PatchTST       & Weather & 336 & 0.2962 & 0.2974 & 0.2948 & 0.0026 \\
iTransformer   & ETTh1   &  96 & 0.3988 & 0.4111 & 0.4165 & 0.0176 \\
iTransformer   & ETTh1   & 336 & 0.5084 & 0.5180 & 0.5163 & 0.0095 \\
iTransformer   & ETTm1   &  96 & 0.3576 & 0.3515 & 0.3455 & 0.0120 \\
iTransformer   & ETTm1   & 336 & 0.4361 & 0.4401 & 0.4349 & 0.0052 \\
iTransformer   & Weather &  96 & 0.1817 & 0.1787 & 0.1787 & 0.0030 \\
iTransformer   & Weather & 336 & 0.2838 & 0.2822 & 0.2812 & 0.0026 \\
\bottomrule
\end{tabular}
\end{table}

%% ============================================================
\section{Full Augmentation Results by Dataset (Group~3)}
\label{app:augmentation}

Table~\ref{tab:fullaugmentation} reports MSE improvement (\%) for
Group~3 (full-data augmentation, $r_{\mathrm{synth}}\in\{0.25,0.5,1.0\}$,
best over ratios) relative to the Group~1 baseline.
Positive values indicate the augmented model beats the real-only baseline.
Architecture-conditional behaviour is consistent across datasets:
TimesNet and iTransformer show net gains on most benchmarks; DLinear
and PatchTST are consistently harmed, with DLinear incurring up to
$-26\%$ on ETTm1 and ETTm2.

\begin{table}[H]
\centering
\caption{Group~3 MSE improvement (\%) over Group~1 baseline per
  (dataset, $H$) pair, best ratio.
  Positive = augmentation improves MSE.  The best-ratio column is an
  oracle summary over the tested ratios, not a deployment-time selection rule.}
\label{tab:fullaugmentation}
\begin{tabular}{@{}lc rrrrr@{}}
\toprule
Dataset & $H$ & DLinear & SegRNN & TimesNet & iTransformer & PatchTST \\
\midrule
\multirow{2}{*}{ETTh1}
  & 96  & $-4.1$ & $-4.4$ & $-1.2$ & $+1.6$ & $-7.3$ \\
  & 336 & $-2.3$ & $-3.1$ & $+7.7$ & $+0.2$ & $-6.7$ \\
\multirow{2}{*}{ETTh2}
  & 96  & $-8.1$ & $+1.5$ & $+9.8$ & $+4.9$ & $-3.5$ \\
  & 336 & $+5.0$ & $+1.9$ & $+11.8$ & $+3.8$ & $-0.4$ \\
\multirow{2}{*}{ETTm1}
  & 96  & $-22.7$ & $-6.3$ & $+2.5$ & $+5.7$ & $-9.0$ \\
  & 336 & $-25.7$ & $-4.7$ & $+1.9$ & $+8.5$ & $-7.6$ \\
\multirow{2}{*}{ETTm2}
  & 96  & $-25.9$ & $-0.6$ & $+7.3$ & $+4.4$ & $-4.8$ \\
  & 336 & $-6.9$  & $-1.8$ & $+2.8$ & $+4.0$ & $-2.3$ \\
\multirow{2}{*}{Weather}
  & 96  & $+3.5$ & $-0.4$ & $+4.2$ & $+1.6$ & $+0.5$ \\
  & 336 & $+1.2$ & $+1.0$ & $+3.4$ & $+0.6$ & $-1.6$ \\
\multirow{2}{*}{Electricity}
  & 96  & $-8.5$ & $-7.7$ & $+1.7$ & $-2.0$ & $-8.2$ \\
  & 336 & $-7.0$ & $-4.0$ & $+1.5$ & $-1.5$ & $-6.9$ \\
\multirow{2}{*}{Traffic}
  & 96  & $-3.2$ & $+1.1$ & $-0.6$ & $+0.6$ & $-2.2$ \\
  & 336 & $-2.4$ & $+1.4$ & $+6.4$ & $-0.6$ & $-3.4$ \\
\midrule
Mean & & $-7.6$ & $-1.9$ & $+4.3$ & $+2.3$ & $-4.6$ \\
\bottomrule
\end{tabular}%
\end{table}

%% ============================================================
\section{Public Experiment Dataset}
\label{app:dataset}

To support reproducibility and future benchmarking, all experiment
results were released as a \href{https://github.com/hugoiscracked/synthetic-ts/blob/main/results/all_results.csv}{single consolidated CSV file}.
The dataset covers all 4,218 runs across Groups~1--9,
with fields for experiment group, model, dataset, prediction horizon,
seed, data mode, synthetic ratio, sparsity, difficulty, bundle,
latent factor probability, annealing strategy, annealing epoch,
MSE, MAE, training time, and timestamp.
This allows independent reanalysis of all findings in this paper
without requiring GPU resources, and provides a foundation for 
meta-learning approaches to hyperparameter selection for synthetic
augmentation.

\end{document}